%% file: neurips_2024.tex
\documentclass{article}

\usepackage{graphicx}
\usepackage{booktabs}
\usepackage[preprint]{neurips_2024}

\usepackage[utf8]{inputenc} % allow utf-8 input
\usepackage[T1]{fontenc}    % use 8-bit T1 fonts
\usepackage{hyperref}       % hyperlinks
\usepackage{url}            % simple URL typesetting
\usepackage{booktabs}       % professional-quality tables
\usepackage{amsfonts}       % blackboard math symbols
\usepackage{nicefrac}       % compact symbols for 1/2, etc.
\usepackage{microtype}      % microtypography
\usepackage{xcolor}         % colors
\usepackage[linesnumbered,ruled,vlined]{algorithm2e}
\usepackage{makecell} % Add this in the preamble
\usepackage{pifont}

\newcommand{\cmark}{\ding{51}}
\newcommand{\xmark}{\ding{55}}
\newcommand{\ie}{\textit{i}.\textit{e}.}

\title{Generalized Open-World Semi-Supervised Object Detection}

\author{
  Garvita Allabadi \quad 
  Ana Lucic \quad
  Siddarth Aananth \quad  
  Tiffany Yang \quad \\
  \textbf{Yu-Xiong Wang} \quad 
  \textbf{Vikram Adve} \\
  University of Illinois at Urbana-Champaign\\
  {\texttt{\{garvita4,alucic2,aananth2,yang244,yxw,vadve\}@illinois.edu}}
  }

\begin{document}

\maketitle

\input{sections/0_abstract}

\input{sections/1_introduction}
\input{sections/2_relatedwork}
\input{sections/3_methodology}
\input{sections/4_experiments}

\input{sections/5_conclusion}

\clearpage
\bibliography{neurips_2024}
\clearpage

\appendix
\input{sections/6_appendix}

\end{document}

%% file: sections/0_abstract.tex
\begin{abstract}
Traditional semi-supervised object detection methods assume a fixed set of object classes (in-distribution or ID classes) during training and deployment, which limits performance in real-world scenarios where unseen classes (out-of-distribution or OOD classes) may appear. In such cases, OOD data is often misclassified as ID, thus harming the ID classes accuracy. Open-set methods address this limitation by filtering OOD data to improve ID performance, thereby limiting the learning process to ID classes. We extend this to a more natural open-world setting, where the OOD classes are not only detected but also incorporated into the learning process. Specifically, we explore two key questions: 1) how to accurately detect OOD samples, and, most importantly, 2) how to effectively learn from the OOD samples in a semi-supervised object detection pipeline without compromising ID accuracy. To address this, we introduce an ensemble-based OOD Explorer for detection and classification, and an adaptable semi-supervised object detection framework that integrates both ID and OOD data. Through extensive evaluation on different open-world scenarios, we demonstrate that our method performs competitively against state-of-the-art OOD detection algorithms and also significantly boosts the semi-supervised learning performance for both ID and OOD classes. 
\end{abstract}

%% file: sections/1_introduction.tex
\section{Introduction}

Object detection based on deep learning techniques is highly dependent on two assumptions (1) availability of large-scale labeled datasets, and (2) a common, fixed set of object classes that appear in both training and unlabeled data, i.e., the ``closed-world'' assumption. While these assumptions may hold in ideal conditions, they often manifest as limitations in real-world situations. Semi-supervised object detection (SSOD) approaches  \citep{sohn2020detection, berthelot2019mixmatch, jeong2019consistency} address the first limitation by leveraging unlabeled data to boost model performance. However, these methods assume the classes in both the training and unlabeled data are sampled from the same distribution, i.e., in-distribution (ID) objects. In practice, the unlabeled data might contain objects that were \textit{unknown} and \textit{unseen} during training, which we call out-of-distribution (OOD) objects. 

Prior work \citep{dhamija2020overlooked, liuopen, miller2021uncertainty} shows that the existing SSOD techniques perform worse in the open-set setting, i.e in the presence of OOD objects. This issue is termed as the `semantic expansion' problem, where OOD objects end up being detected as ID objects and incorrectly used as pseudo-labels for unlabeled data. 
\cite{liuopen} addresses the open-set semi-supervised object detection problem, by proposing an offline OOD detector to filter out (reject) any OOD data, to improve the performance of ID classes. These methods, however, disregard any novel OOD data, thereby limiting the learning process to a predefined set of classes.

\begin{figure}
    \centering
    \includegraphics[width=0.95\textwidth]{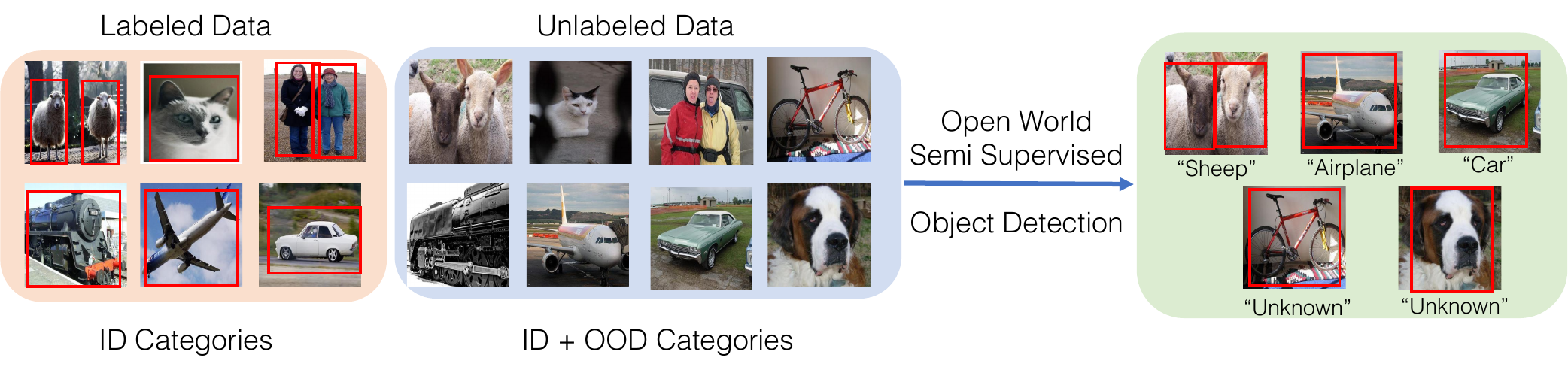}
    \caption{In our task of generalized open-world semi-supervised object detection (OWSSD), the model observes a labeled dataset with ID data and an unlabeled dataset which can contain both ID and OOD data. The objective is to train a detection model that is able to (1) localize and classify instances belonging to seen classes; and (2) localize instances not belonging to seen classes and group them into a new `unknown' class.}
    \label{fig:owssd-problem}
\end{figure}

In this work, we propose an extension of the open-set to a more generalizable open-world setting that not only improves the performance for ID classes but also discovers and includes OOD classes in the learning process. Specifically, in the generalized open-world setting, we are given a small labeled dataset that includes ID categories of data and a large unlabeled dataset that contains both ID and OOD categories. The aim is to use both the labeled and unlabeled data to improve the performance of ID categories and simultaneously discover OOD data as `unknown' and adapt the semi-supervised object detection pipeline (Figure \ref{fig:owssd-problem}).

To achieve this, we propose an \textit{OOD Explorer} which is designed to address two primary tasks: the classification of objects as OOD or ID and the localization of novel OOD objects. For the OOD classification task, we propose a simple yet effective OOD detector that leverages an ensemble of lightweight auto-encoder networks trained \textit{only} on ID data. For OOD localization, we systematically analyze multiple unsupervised, class-agnostic object detection techniques, as they align with the inherently open-world nature of the problem. The second component is an \textit{OOD-aware semi-supervised learning framework}, structured into two learning stages. The first stage leverages labeled data, while the second stage incorporates both labeled and unlabeled data, with the help of the OOD Explorer that introduces OOD data.

In our experiments, we examine different open-world scenarios and evaluate the performance of (1) the proposed OOD Explorer on both OOD classification and localization metrics (2) OOD-aware semi supervised object detection. The results show that our method performs competitively against state-of-the-art OOD detection algorithms and that the proposed OOD-aware semi-supervised learning method significantly improves the robustness of ID objects classification and identification through its ability to detect OOD objects and integrate them into the model and learn from them.
Our contributions are summarized as follows:
\begin{itemize}
     \item We address the problem of generalized open-world semi-supervised object detection problem, by improving the  performance on \textit{both} ID and OOD classes.
     \item We propose a novel \textit{OOD Explorer} for handling novel OOD data, capable of both classification and localization) (Section~\ref{sec:methodology:ood-explorer}). In combination with the OOD Explorer, we develop an OOD-aware semi-supervised object detection algorithm capable of detecting and labeling objects of both ID and OOD classes (Section~\ref{sec:methodology:ssl-pipeline}).
    \item We demonstrate that our proposed method is able to detect OOD classes while substantially improving the robustness of ID class detection. (Section~\ref{sec:expts}).
\end{itemize}

%% file: sections/2_relatedwork.tex
\section{Related Work}
\label{sec:related}

\begin{table}[t]
\small
\centering
\caption{Comparison of our method with related settings}
\begin{tabular}{cccc}
\toprule
Setting & Detect Novel OOD Data & Semi-Supervised & Learns from Novel OOD Data \\
\midrule
SSOD & \xmark & \cmark & \xmark \\ 
Open-World OD & \cmark & \xmark & \cmark \\
Open-Set SSOD & \cmark & \cmark & \xmark \\ \midrule
\textbf{Our Method} & \cmark & \cmark & \cmark \\
\bottomrule
\end{tabular}
\label{tab:comparison}
\end{table}

\paragraph{Semi-Supervised Object Detection.} Semi-supervised object detection (SSOD) approaches have become popular to reduce the need for labeling \cite{sohn2020detection, berthelot2019mixmatch, jeong2019consistency}. Pseudo-labeling based methods such as FlexMatch \cite{zhang2021flexmatch}, TSSDL \cite{shi2018transductive}, and others \cite{iscen2019label, luo2018smooth, yan2019semi, liu2021unbiased, xu2021end}, first train a teacher model using only labeled data and then use that model to create pseudo-labels for unlabeled images. The pseudo-labels are then used along with the original labeled data to train a student model. On the other hand, consistency regularization approaches such as \cite{sajjadi2016regularization, laine2017temporal, tarvainen2017mean, liu2021certainty, luo2018smooth, jeong2019consistency, iscen2019label, liu2021unbiased, xu2021end}, aim to minimize a consistency loss between differently augmented versions of an image. All of these semi-supervised learning approaches assume a ``closed-world'' setting with a fixed set of classes in both training and testing, which is not a valid assumption in real-world applications.

\paragraph{Open-World Object Detection.} Open-world object detection enables the detection of novel objects by incrementally adding novel object classes to the set of known classes. Previous work \cite{kim2022learning, kuo2015deepbox, o2015learning, wang2020leads, Maaz2022Multimodal} has studied different methods of object proposals for novel objects by attempting to remove the notion of class (all objects are regarded the same). ORE \cite{joseph2021towards} is the first to propose an open-world object detector that identifies novel classes as ‘unknown’ and proceeds to learn the unknown classes once the labels become available. \cite{han2022expanding} aims to identify unknown objects by separating high/low-density regions in the latent space. Both these approaches work in a fully-supervised setting. Our setup goes a step further and situates the open-world problem in the context of semi-supervised learning, with limited amounts of labeled ID data \textit{only}, that more closely resembles the real-world settings. 

\paragraph{Unsupervised Object Localization.} Recently proposed methods such as CutLER \cite{wang2023cut}, FreeSolo \cite{wang2022freesolo}, LOST \cite{LOST}, and MOST \cite{rambhatla2023most} propose to localize objects in an unsupervised manner, either by segmentation masks or bounding boxes. Some of these \cite{wang2023cut, LOST, rambhatla2023most} use features from self-supervised trained transformers to localize objects in the scene. In our work, we evaluate the capabilities of such methods for localizing OOD objects, as they present open-world capabilities. Based on our evaluation (\ref{sec:expts:ablation}), we use CutLER as part of the OOD Explorer to localize OOD classes. Section \ref{sec:expts} provides the details of our evaluation. 

\paragraph{Open-Set/Open-world Semi-Supervised Object Detection.}
The open-set semi-supervised object detection problem \cite{liuopen} addressed some of the limitation of the above mentioned work. Furthermore, they address like the performance of ID classes in the presence of OOD data, but they do not learn from it or improve OOD performance. They propose an offline OOD detector to filter out OOD data, thus limiting the risk of ID performance in the presence of OOD data. In contrast, our approach \textit{both} improves performance for ID classes \textit{as well as} OOD classes, i.e., our proposed framework solves a strictly stronger problem. Specifically speaking, \cite{liuopen} solves for identifying novel classes and filters it out, but does not re-introduce the classes back into the training pipeline in order to be able to learn its features. \cite{mullappilly2024semi} addresses some of the limitations of the previous mentioned methods by extending the problem to a semi-supervised setting. However, their problem setting is similar to an incremental learning setting, access to unknown class labels is provided in subsequent tasks. Our generalized setting, on the other hand, does not require access to any unknown class labels. 

%% file: sections/3_methodology.tex
\section{Generalized Open-World Semi-Supervised Object Detection}
\label{sec:method}

\begin{figure}[t]
\centering
\includegraphics[width=0.95\textwidth]{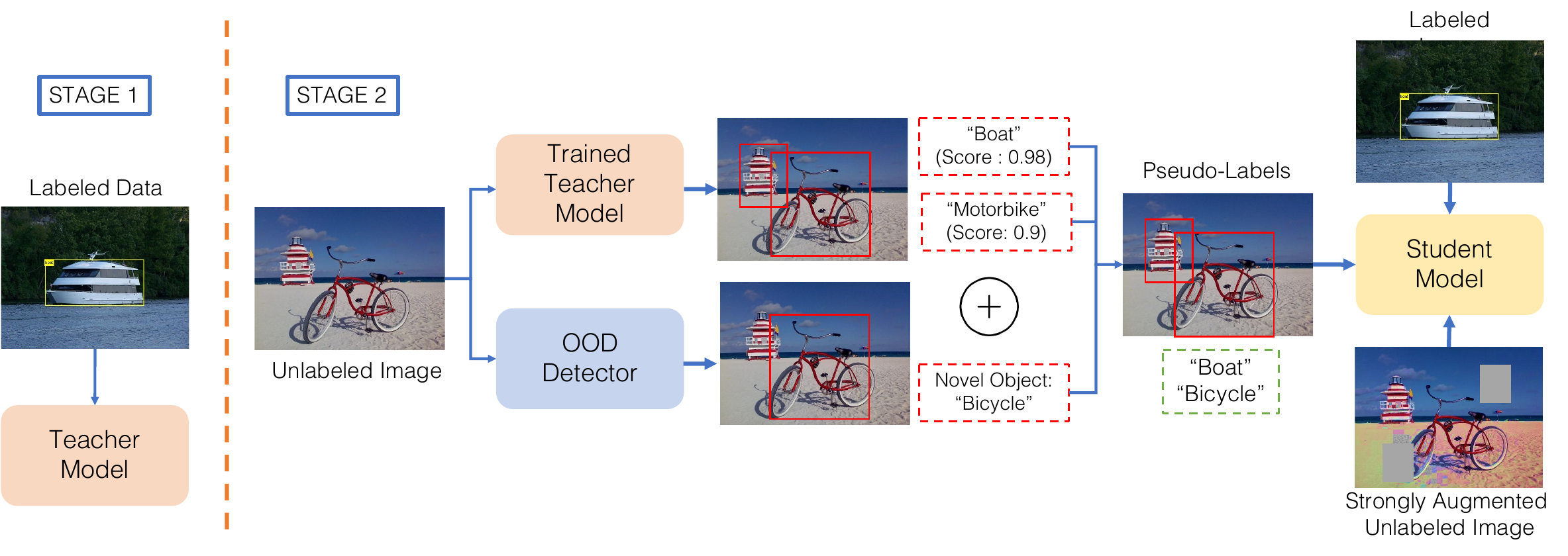}
\caption{A conceptual overview of our \textbf{OOD-aware semi-supervised learning} framework. In the first stage, a \textbf{Teacher Model} is trained from the available labeled images with ID categories. In the second stage, unlabeled images that could contain ID and OOD categories are passed through (1) the trained \textbf{Teacher Model} to generate pseudo-labels for the ID categories that the Teacher model was trained on; (2) an \textbf{OOD Detector} to discover any novel OOD categories present in the unlabeled data. The pseudo-labels generated by the two modules are then fused such that in cases of conflict (high overlap), the novel OOD labels get preference. This helps eliminate false negatives that could be introduced when an OOD category is wrongly classified as an ID category. Finally, a \textbf{Student model} is trained jointly on the labeled and unlabeled data to minimize the total loss.}
\label{fig-owssd-method}
\end{figure}

We propose an integrated framework for \textit{Generalized Open-World Semi-Supervised Object Detection} consisting of an Ensemble-Based OOD Explorer (Section \ref{sec:methodology:ood-explorer}) and an OOD aware semi-supervised pipeline (Section \ref{sec:methodology:ood-explorer}). The OOD Explorer plays a crucial role in identifying OOD objects, encompassing both localization and classification into ID or `'unknown' class. The OOD-aware semi-supervised learning pipeline ( Section \ref{sec:methodology:ssl-pipeline}) ensures that the model assimilates knowledge from both ID and OOD data without risking the forgetting of previously learned ID classes. This integrated approach aims to enhance the adaptability and robustness of object detection models in the open-world context. Figure \ref{fig-owssd-method} provides a visual summary of our proposed framework.

\subsection{Problem Formulation}\label{sec:OWSSD:Problem} 

For our task of generalized open-world semi-supervised object detection, we are given a small labeled dataset \textit{$D_l = \{(x_1,y_1),(x_2,y_2), \ldots, (x_n,y_n)\}$} and a large unlabeled dataset \textit{\textit{$D_u = \{u_1, u_2, \ldots, u_m\}$}}, where $\{x_i\}$ and $\{u_i\}$ are input images, and $\{y_i\}$ are annotations. \textit{$D_l$} consists of a set of ID categories denoted by \textit{$C_\mathrm{id}$} (\ie, \textit{\textit{$C_{l}$} = \textit{$C_\mathrm{id}$}}), and \textit{$D_u$} consists of both \textit{$C_\mathrm{id}$} and a set of OOD categories denoted by \textit{$C_\mathrm{ood}$} (\textit{\textit{i.e., $C_{u}$} = \textit{$C_\mathrm{id}$} $\cup$ \textit{$C_\mathrm{ood}$}}). Each annotation \textit{$y_i$} in \textit{$D_l$} includes a set of object bounding box coordinates and the corresponding class labels from \textit{$C_\mathrm{id}$}.

Our objective is to train a detection model on \textit{$D_l$} and \textit{$D_u$} jointly that is able to (1) localize and classify instances belonging to one of the ID categories \textit{$C_\mathrm{id}$}; and (2) identify instances belonging to \textit{$C_\mathrm{ood}$} and localizing them. Note that the model does not have access to any \textit{$C_\mathrm{ood}$} bounding box coordinates or labels. The evaluation is performed on a held-out dataset \textit{$D_v$} that consists of both objects from \textit{$C_\mathrm{id}$} and \textit{$C_\mathrm{ood}$} categories (\ie, \textit{\textit{$C_{v}$} = \textit{$C_\mathrm{id}$} $\cup$ \textit{$C_\mathrm{ood}$}}).

\subsection{Ensemble-Based OOD Explorer}
\label{sec:methodology:ood-explorer}

In the context of object detection, the integration of OOD data involves two primary tasks: OOD Classification, which distinguishes OOD data from ID data, and OOD Localization, which concurrently provides the precise location of OOD objects within the image. 

\begin{figure}
    \centering
    \begin{minipage}[b]{0.49\textwidth}
        \centering
        \includegraphics[width=0.95\textwidth]{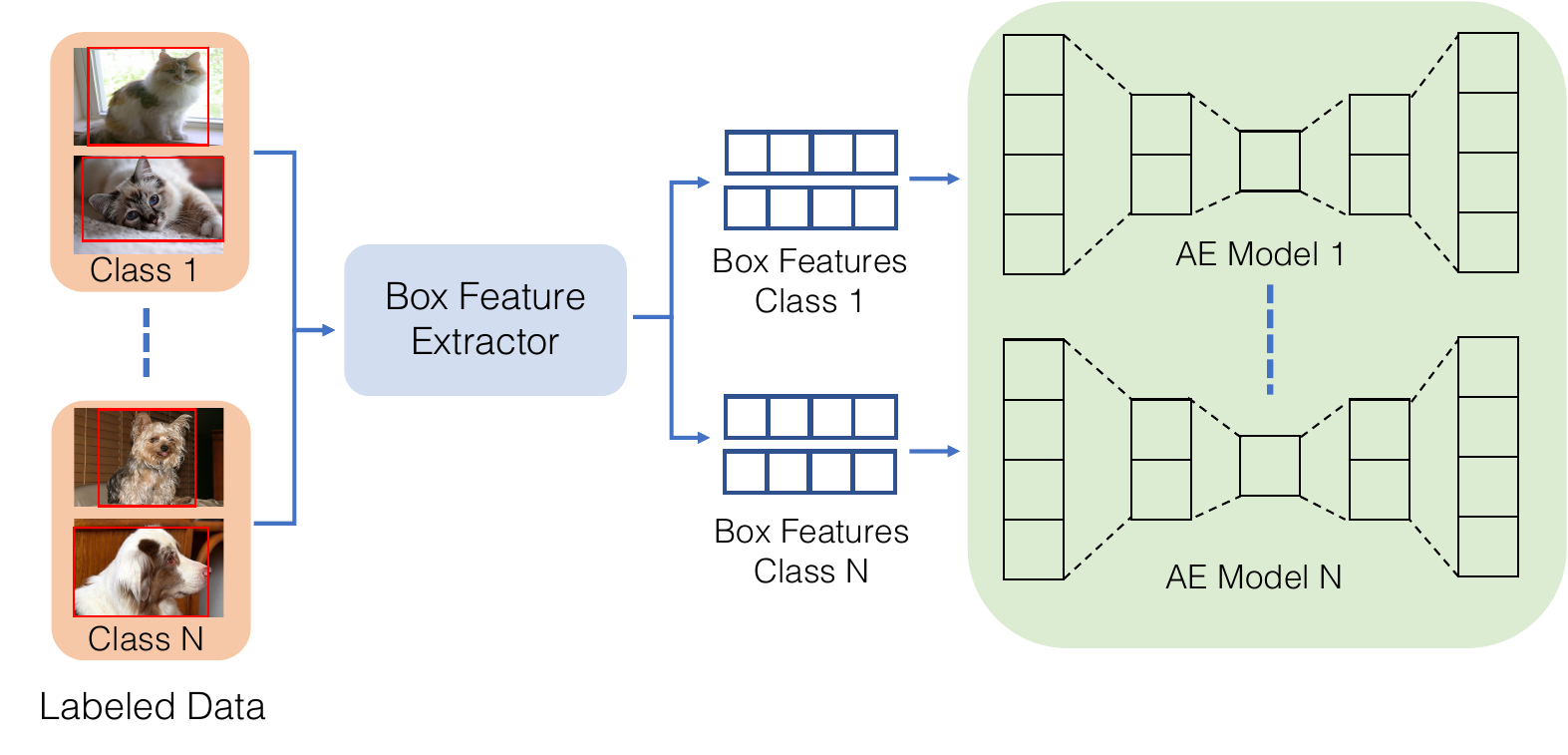} 
        \caption{Training Stage}
    \end{minipage}\hfill
    \begin{minipage}[b]{0.49\textwidth}
        \centering
        \includegraphics[width=0.95\textwidth]{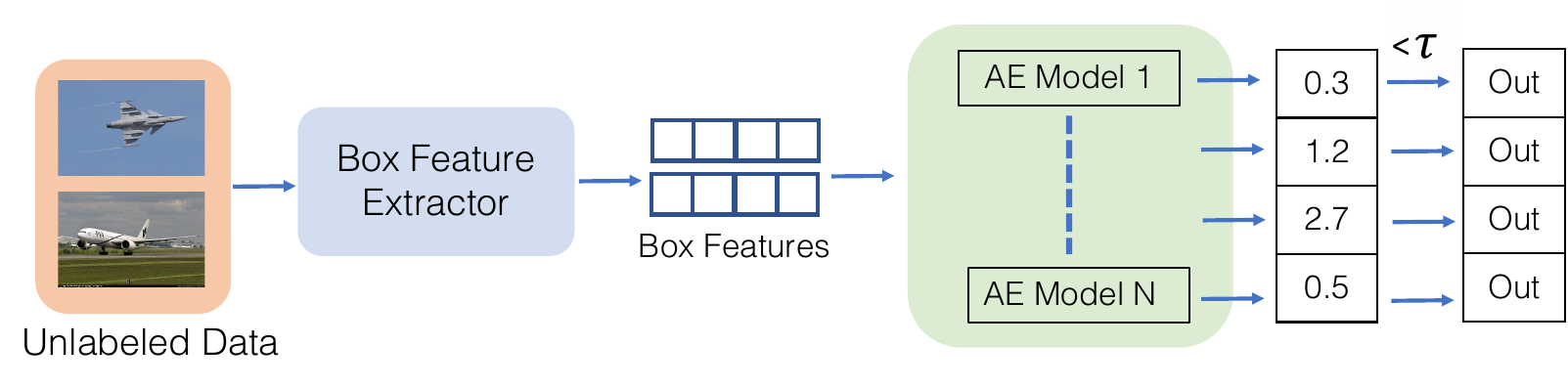}
        \caption{Testing Stage}
    \end{minipage}
    \caption{Proposed Ensemble Based OOD Detector. In the training stage, ground truth proposals from the labeled data are passed through the feature extractor, the output of which is used to train our OOD detector to learn the representation of ID classes. In the OOD prediction stage, unlabeled images are passed through class-agnostic and open-world friendly proposal generators to generate proposals that are then predicted as ID/OOD by the trained OOD detector.}
    \label{fig:ensemble}
\end{figure}

\textbf{OOD Classification}. We propose an ensemble model that is trained using labeled ID data \textit{only} and without OOD samples. Our ensemble model comprises of multiple auto-encoder networks, each trained on samples from a specific ID class. Each model within the ensemble is dedicated to learning a lower-dimensional representation (encoding) for its respective category. 
An autoencoder network uses an encoder ($z=E_{\phi }(x), E_{\phi }:\mathcal {X}\rightarrow \mathcal {Z}$) to compress its input data and a decoder ($\hat{x}=D_{\theta }(z), D_{\theta }:\mathcal {Z}\rightarrow \mathcal {X}$)to reconstruct the original input. 

The autoencoder learns by minimizing the reconstruction error $R$, which gives a measure of how well the output was reconstructed compared to the input. $R$ is estimated with a dissimilarity function $d$, such that $d(x,\hat{x})$ measures how much $x$ differs from $\hat{x}$ ($R = d(x,D_{\theta }(E_{\phi }(x)))$)

When an auto-encoder model is trained only with a specific category of ID data, the reconstruction error, $R$, for samples of that category is low. At test time, when confronted with an OOD sample, the model is unable to accurately reconstruct the input, resulting in a higher reconstruction error. If this error value is less than a threshold value ($\mu$), then the sample belongs to the corresponding ID class which the model was trained on. This process is repeated for all auto-encoder models in the ensemble. Each model votes if the sample belongs to its class based on the observed $R$. A sample is deemed OOD if none of the models ``claim'' it. This simple heuristic illustrates how the decision boundary between ID and OOD samples is established (Figure \ref{fig:ensemble} shows this process).

This threshold is calibrated during training by observing the reconstruction error of ID categories. Since each model is responsible only for its own category, all other categories function as pseudo-OOD data for determining $\mu$. Finally, given that a single image may contain multiple objects of diverse categories, we leverage the box-level features of each object present in the image (Refer to the appendix for details on the training and test procedure).

As seen above, training an autoencoder network is inherently unsupervised as it does not require any information about the class label. By introducing an ensemble of such networks trained on individual categories, we introduce a specialization for each model. \emph{Our key insight is that this introduced specialization is a useful and necessary requirement for OOD detection with very limited labeled data.} When an autoencoder model is trained using the objects of a single category, the model learns the most salient and informative characteristics for that category. Training such an ensemble of models ensures that a model makes confident predictions when encountered with objects from the same ID category and results in a high reconstruction error when encountered with an object different from the one it was trained on. Section \ref{sec:expts} describes the variety of experiments we conduct which demonstrate that an ensemble model performs better compared to not only a common autoencoder model trained on all ID classes, but also a variety of state-of-the-art OOD detection algorithms. 

\textbf{OOD Localization}. In standard two-stage detectors like Faster R-CNN, object localization is achieved by using a Region Proposal Network (RPN) to generate object proposals, a selection of locations likely to contain objects. This network, however, is trained on a fixed set of ID classes and thus fails to generalize to novel OOD classes.

To address this challenge, we employ CutLER \citep{wang2023cut}, an unsupervised method that generates class-agnostic proposals. This characteristic proves advantageous in our open-world setting, as the proposals are not confined to in-distribution (ID) classes. We utilize the models pre-trained on unsupervised ImageNet, including both Cascade Mask RCNN and Mask RCNN. Subsequently, we filter the proposals based on a confidence score.

Additionally, we also analyzed two other localization methods for OOD localization, namely OLN \citep{kim2022learning} and MOST \citep{rambhatla2023most}. CutLER and MOST use the features extracted from a transformer \citep{vaswani2017attention} network trained with with a self-supervised manner proposed in DINO \citep{caron2021emerging} to localize multiple objects in an image. OLN \citep{kim2022learning}, on the other hand, estimates the objectness of a candidate region by relying on geometric cues such as location and shape of an object, regardless of its category. We show the performance of these methods when incorporated in our OOD Explorer and SSL pipelines in Section \ref{sec:expts:ablation}. 

\subsection{OOD-Aware Semi-Supervised Learning}
\label{sec:methodology:ssl-pipeline}

We now describe how the OOD samples identified by the OOD Explorer are introduced for training along with the ID samples, in our all {\em OOD-aware} semi-supervised learning framework. 

We adopt the Teacher-Student paradigm and use a two-stage training process (Fig.~\ref{fig-owssd-method}). In the first stage, we use the labeled data to train a \textbf{Teacher model}. This model then operates on weakly-augmented unlabeled images to generate bounding boxes and class predictions for the \textit{ID} classes. A subset of confident predictions that are higher than a pre-determined threshold are treated as pseudo-labels. 

In the open-world setting, such pseudo-labels could be highly noisy -- a novel OOD sample might be wrongly classified as an ID category. Therefore, we filter out the predictions made by the  Teacher model that have an Intersection over Union (IoU) score greater than a threshold $\tau$ with the OOD pseudo labels. 

In the second stage, a \textbf{Student model} is trained using both the labeled data and strongly augmented unlabeled data and the consistency regularization paradigm is used which enforces a model to output the same prediction for an unlabeled sample even after augmentation. The final loss is computed for the labeled samples and unlabeled samples (which includes both ID and OOD categories): $\mathcal{L} =~ \mathcal{L}_s + \lambda \mathcal{L}_{u} $

The supervised loss $\mathcal{L}_s$ is the standard loss for Faster RCNN network comprising of the the RPN classification loss (${L}_{cls}^{rpn}$), the RPN regression loss (${L}_{reg}^{rpn}$), the ROI classification loss (${L}_{cls}^{roi}$), and the ROI regression loss (${L}_{reg}^{roi}$). This loss is computed on the model prediction compared to the ground truth labels. In contrast, the unsupervised loss $\mathcal{L}_u$ is computed on the model prediction on a strongly augmented image compared to the pseudo-label, also consists of the four loss terms. $\lambda$ is a hyper-parameter that controls contribution of the unsupervised loss to the total loss. The details of data augmentation and hyper-parameter settings are described in Sec.~\ref{sec:expts}.

Note that since we are introducing a new OOD class to the model, instead of using the weights provided by the Teacher model, the Student model is retrained from scratch to alleviate the \textit{catastrophic forgetting} issue observed in incremental learning scenarios \citep{li2017learning}.

%% file: sections/4_experiments.tex
\begin{table*}[t]
\begin{center}
\begin{tabular}{l|l|ccc|ccc}
\toprule
& & \multicolumn{3}{c|}{AP$^{50}$$\uparrow$} & \multicolumn{3}{c}{AR$\uparrow$} \\
\midrule
Method & Localization & All & ID & OOD & All & ID & OOD \\
\midrule
Labeled Only & RPN  & 57.5 & 61.3 & - & 42.8 & 45.6 & - \\
SSL & RPN & 60.3 & 64.3 & -  & 46.6 & 49.7 & - \\
SSL + OOD \scriptsize{Filtering} & RPN & 60.4 & 64.4 & - & 46.8 & 49.9 & - \\
\midrule
SSL + OOD \scriptsize{Expansion} & RPN + CutLER & 61.0 & 64.4 & 10.2 & 48.2 & 50.0 & 21.2 \\
SSL + OOD \scriptsize{Expansion} & RPN + CutLER* & 61.2 & 64.7 & 8.8 & 48.4 & 50.1 & 21.8 \\
\bottomrule
\end{tabular}
\end{center}
\caption{Generalized Open-world semi-supervised object detection results on the VOC-15 data split. We examine different open-world settings. The first three settings - Labeled only (Supervised), Semi-supervised (SSL) and OOD Filtering are focused on improving the performance for the original ID classes in the presence of OOD objects. The OOD Expansion setting measures model performance when (unlabeled) OOD objects are also included. 
}
\label{tab:voc-15-all}
\end{table*}

\section{Experiments}
\label{sec:expts}

\subsection{Experimental Setting and Datasets}

\noindent\textbf{Datasets.} 
We evaluate OWSSD on the PASCAL VOC ~\citep{c3} object detection datasets. The trainval sets of VOC07 and VOC12 consist of 5,011 and 11,540 images respectively, both from 20 object categories. The VOC07 test set consists of 4,952 images from 20 object categories.  We evaluate the open world semi supervised object detection by creating a VOC-15 set that samples the first 15 categories from the VOC07 trainval set as the ID classes.  The VOC12 trainval set, with the original 20 categories (15 ID classes and 5 OOD classes), is used as the unlabeled dataset. The model performance is evaluated on the VOC07 test set. 

\noindent\textbf{Evaluation Metrics.} We report two kinds of metrics: (1) Object-detection metrics for end-to-end evaluation: We report the the average precision at IoU=0.50 (AP$^{50}$) and average recall (AR) for all, ID, and OOD object categories. (2) Open-world metrics: To demonstrate the effectiveness of OOD detection, we report the $F_1$ score, False Positive Rate (FPR) and Area Under the ROC curve (AUROC) metrics.  

\subsection{Main Results}
\label{sec:expts:main}

\begin{table}
\begin{minipage}[t]{0.45\textwidth}
\begin{center}
\caption{Evaluation of Unsupervised Localization Methods for the OOD Explorer.}
\begin{tabular}{l|cccc}
\toprule
Method & AP$^{50}$$\uparrow$ & AR$\uparrow$\\
\midrule
 OLN & 3.5 & 14.4  \\
 MOST & 5.2  & 16.9 \\ 
 Oracle & 36.8 & 45.1 \\ 
\midrule
CutLER & 10.2 & 21.2 \\
CutLER* & 8.8 & 21.8 \\
\bottomrule
\end{tabular}
\end{center}
\label{tab:localization}
\hfill
\end{minipage}
\hspace{0.5cm}
\begin{minipage}[t]{0.45\textwidth}
\caption{Evaluation of OOD detection methods for object detection.}
\begin{tabular}{l|cccc}
\toprule
Method & $F_1$ $\uparrow$ & FPR $\downarrow$ & AUROC $\uparrow$\\
\midrule
LOF & 0.33 & 0.44 & 0.68 \\
IF & 0.37 & 0.32 & 0.70   \\
OneSVM & 0.33 & 0.49 & 0.68 \\
KNN& 0.35 & 0.44 & 0.70  \\
\midrule
Ours & \textbf{0.46} & \textbf{0.14} & \textbf{0.72} \\
\bottomrule
\end{tabular}
\label{tab:ood-detection-other-methods}
\end{minipage}
\end{table}

\noindent\textbf{Experiments with OOD Expansion:}
The aim for these experiments is to evaluate the efficacy of our proposed technique for open world semi supervised object detection. We examine different open-world settings. The Labeled only, Semi-supervised (SSL) and OOD Filtering settings are focused on improving the performance for the original ID classes in the presence of OOD objects. The OOD Expansion setting measures model performance when (unlabeled) OOD objects are also included in the training process. Our results are presented in Table \ref{tab:voc-15-all}.

We find that introducing OOD filtering through our ensemble OOD detector, combined with SSL, enhances the performance of SSL across ID and All classes, compared to the Label Only and SSL baselines. 
Additionally, when adapting SSL for OOD Expansion using our proposed \textit{OOD Explorer}, we observe substantial improvements, including an increase of +5.3 mean Average Precision (mAP) for all classes and +5.4 mAP for ID classes, compared with using labeled images only, and +0.7 and +0.5 compared with the baseline SSL. Furthermore, our method, in conjunction with CutLER for OOD Localization, demonstrates increased capabilities in detecting OOD objects, achieving an AP$^{50}$ of 10.2 and an Average Recall (AR) of 21.2.
Cutler ~\citep{wang2023cut} uses the MaskRCNN, and using a variant of CutLER with Cascade Mask RCNN (which we label CutLER*) further improves performance slightly for ID objects but hurts slightly for OOD objects.

\noindent\textbf{Comparison of OOD Detection Methods:}
We compare the OOD detection capabilities of our proposed ensemble method with other OOD detection methods, namely Local Outlier Factor (LOF) ~\citep{10.1145/342009.335388}, Isolation Forest (IF) \citep{4781136}, One Class Support Vector Machine (OneSVM)~\citep{10.1162/089976601750264965} and K-Nearest Neighbors (KNN). Table \ref{tab:ood-detection-other-methods} summarizes our findings. OWSSD's OOD detector features a significantly higher $F_1$ score compared to existing OOD detection methods with the same amount of limited labeled data. Also, the OWSSD OOD detector results in the highest AUROC score that summarizes the ability of the model to distinguish between the seen and unseen classes. The False Positive Rate reflects the rate of false positive OOD instances: the lower the rate the better performance of the model. The FPR rate for OWSSD OOD is the lowest, which explains why our technique is able to outperform the other methods.

\subsection{Ablation Studies}
\label{sec:expts:ablation}

\noindent\textbf{Evaluation of OOD Localization methods}
We evaluated 3 different localization methods for our \textit{OOD Explorer}, namely CutLER \citep{wang2023cut}, MOST \citep{rambhatla2023most} and OLN \citep{kim2022learning}. We also compared the performance against an Oracle that generates perfect proposals for OOD classes, to establish an upper bound. Based on the proposals provided by each of these methods, we use our ensemble based OOD detector to identify the OOD proposals. The performance for OOD classes is reported after training the student model. 

As seen in Table \ref{tab:localization}, CutLER outperforms both OLN and MOST for OOD object detection. Nevertheless, considering the inherent challenges in localizing OOD data without any labels in the context of open-world learning, there remains potential for enhancement, as evidenced by the much better performance of the oracle proposal generator. These results show that accurate localization is a key bottleneck we must address for further performance improvements on open-world semi-supervised object detection.

\noindent\textbf{Impact of different error thresholds for OOD detector:}
OWSSD's OOD detectors consisting of an ensemble of autoencoder networks categorizes data into ID or OOD classes based on a threshold value $\mu$. We study how the threshold value impacts the OOD detector's performance. The lower reconstruction error indicates that one of the autoencoder models has been able to reconstruct the object and that object class has been claimed. If, however, none of the classes can effectively reconstruct the object featured in the bounding box, the voting mechanism of separate encoders would result in a higher reconstruction error. We tested a series of threshold ($\mu$) values and show results for the following: 0.05, 0.1, 0.2. As Table 4 indicates, the best balance between precision and recall, as reflected through the $F_1$ score, is achieved with the 0.1 threshold. 

\begin{table}
\begin{minipage}[t]{0.45\textwidth}
\begin{center}
\caption{Choice of reconstruction error threshold for ID vs OOD categorization}
\begin{tabular}{l|cccc}
\toprule
Threshold & $F_1$ $\uparrow$ & FPR $\downarrow$ & AUROC $\uparrow$\\
\midrule
 0.05 & 0.40  & 0.35 & 0.74  \\
 0.2 & 0.29 & 0.003  & 0.59 \\ 
\midrule
Ours 0.1 & 0.46 & 0.14 & 0.72\\
\bottomrule
\end{tabular}
\end{center}
\label{tab:threshold}
\end{minipage}
\hspace{0.5cm}
\begin{minipage}[t]{0.45\textwidth}
\begin{center}
\caption{Impact of using an Ensemble compared to a common AE model.}
\begin{tabular}{l|ccc}
\toprule
Method & $F_1$ $\uparrow$ & FPR $\downarrow$ & AUROC $\uparrow$\\
\midrule
Common AE & 0.42 & 0.09 & 0.67\\
Ours & 0.46 & 0.14 & 0.72 \\
\bottomrule
\end{tabular}
\end{center}
\label{design}
\end{minipage}
\end{table}

\noindent\textbf{Evaluating ensemble approach in the OOD detector.} OWSSD uses an ensemble of autoencoder networks for OOD detection. We compared the performance of OWSSD's OOD detector to a common autoencoder model trained for all ID categories with limited labeled data. Table \ref{design} shows the results. The ensemble method results in a better recall and $F_1$ score for OOD detection compared to a common model trained on all labeled ID data, showing that it is able to distinguish OOD classes effectively. 
As noted previously, the ensemble approach enables each autoencoder to have high accuracy in distinguishing the specific ID class it was trained on, which leads to better discrimination of ID vs OOD objects. (The small increase in False Positive Rate did not affect the F1 and AUROC score advantage that the Ensemble AE method has over the Single AE.)

%% file: sections/5_conclusion.tex
\section{Conclusion}
\label{sec:conclusion}

We introduce an generalized OOD-aware semi-supervised object detection approach, based on a Teacher-Student paradigm, which includes both ID and OOD data. We also propose, an OOD Explorer to achieve two goals: 1) effective separation of OOD objects from ID objects and 2) localization of OOD objects. Our results show that this method significantly improves the robustness of semi-supervised object detection methods for ID objects through its capability to not only detect OOD objects but also integrate them into the training process and learn from them.   

\noindent\textbf{Limitations and future work.} While our method advances the state of the art, there is much room for improvement with respect to the localization and classification of OOD objects.
Most importantly, new techniques are needed to localize OOD objects more precisely. Our future work will include a repeated and continual testing of the proposed approach with a varying number of new classes. New approaches such as continuous, in-field learning and the incorporation of domain understanding into current data-driven models may hold promise in achieving these goals.

%% file: sections/6_appendix.tex
\clearpage

\title{Appendix}

\maketitle

\section{Implementation Details}

We use Faster R-CNN equipped with a Feature Pyramid Network and a ResNet-50 backbone as our object detector. The backbone is initialized with ImageNet pre-trained weights in both stages of training. The implementation is based on the torchvision library \citep{c1} from PyTorch \citep{c2}.  For the open-world settings, we use $\mu=0.1$ as the threshold for VOC-15 data split. The autoencoder model architecture in this paper uses one hidden layer in the encoder and decoder parts of the network. A Feature Pyramid Network vector, 1,024 in length, is used as input to the model that is then compressed to 256 and eventually to 64 in the bottleneck layer. For the semi-supervised settings, we use $\tau=0.9$ as the pseudo-labeling threshold, $\lambda=2$ as the unlabeled loss weight. Our data augmentation strategy consists of RandomHorizontalFlip for the first stage of training, and a combination of ScaleJitter, FixedSizeCrop, and RandomHorizontalFlip for the second stage of training. For Cutler, we use the ImageNet pre-trained models using Cascade Mask R-CNN and Mask R-CNN as detectors.

\section{OOD detection algorithms}

\begin{minipage}[t]{0.46\textwidth}
\begin{algorithm}[H]
\caption{OOD Detector Training Algorithm}
\KwData{$D_l$: Images and annotations consisting of class labels and bounding boxes for objects of \textit{$C_\mathrm{id}$} classes}
\KwResult{Ensemble of auto-encoder trained models}
\ForEach {c in \textit{$C_\mathrm{id}$}}{
\ForEach {bb in $D_l$ where class = $c$}
{
$x = extract\_features(bb)$\\
$z = E_{\phi }(x)$\\
$\hat{x}=D_{\theta }(z)$ \\
$ L(x, \hat{x}) = \frac{1}{N} \sum_{i=1}^{N}||x_i-\hat{x_i}||^2$ 
}
}
\end{algorithm}
\end{minipage}
\hfill
\begin{minipage}[t]{0.46\textwidth}
\begin{algorithm}[H]
\caption{OOD Detector Testing Algorithm}
\KwData{$D_u$: Images and class-agnostic proposals for objects of \textit{$C_\mathrm{id}$} and \textit{$C_\mathrm{ood}$} classes;\\
Ens: Ensemble of autoencoder trained models}
\KwResult{List of OOD proposals}
\ForEach {bb in $D_u$ where class = $c$}
{
\ForEach {model in Ens}{
$X = extract\_features(bb)$\\
$z = E_{\phi }(x)$\\
$\hat{x}=D_{\theta }(z)$ \\
$ L_(x, \hat{x}) = \frac{1}{N} \sum_{i=1}^{N}||x_i-\hat{x_i}||^2$ \\
\uIf{$ L_(x, \hat{x}) < \mu$}{
     id = \textbf{True} \\
  }
}
\uIf{not id}{
    ood\_list.add(bb)
  }
}
\textbf{return} {ood\_list}
\end{algorithm}
\end{minipage}

\section{Additional ablation studies}
\subsection{Impact of choices in our OOD-aware semi-supervised pipeline.}
In our OOD-aware semi-supervised pipeline, if generated pseudo-labels for ID classes (i.e. high quality predictions made by the Teacher model) have an Intersection over Union (IoU) score greater than a threshold with the OOD pseudo labels, they are filtered out from the final list of pseudo labels. In this study, we examine the impact of not filtering out the overlapping pseudo-labels. As seen in table \ref{table-abl-ssl}, when OOD pseudo-labels and ID pseudo-labels are merged with filtering based on their IoU scores, the overall performance for both ID and OOD classes increases, as no false positives are introduced.

\begin{table}
\begin{center}
\caption{Impact of the OOD-awareness in the semi-supervised object detection pipeline. Our approach results in increased performance by reducing the false positive pseudo labels.}
\begin{tabular}{l|ccc}
\toprule
mAP & All & ID & OOD  \\
\midrule
Without adding OOD data & 60.3 & 64.3 & -  \\
With OOD data, without filtering  & 63.4 & 65.8 & 27.1  \\
\midrule
With OOD data + filtering & 64.3 & 66.1 & 36.8 \\
\bottomrule
\end{tabular}
\end{center}
\label{table-abl-ssl}
\end{table}

\subsection{Impact of number of layers in the autoencoders in the OOD detector}

We conducted an ablation study on the architecture of the OOD detector. We increased the number of layers from 3 to 5, including the input and output layers. In our Ensemble autoencoder network, we start with 1,024 features that are compressed to 256 and then to 64 in the bottleneck layer. In the architecture with 5 layers, we start with the 1,024 features that are compressed to 512, 256, 128, and 64 in the bottleneck layer. The decoder part is symmetrical to the encoder part. In both architectures, each layer is followed by the Rectified Linear Activation (ReLU) activation function, the number of epochs is held constant at 30, the mean squared error is used as the loss function, and the learning rate is set at 0.001. As Table \ref{tab:design} indicates, we see a slight improvement in the AUROC score with the network with more layers (0.74 versus 0.72). The $F_1$ score, however, has seen a slight decrease (0.43 compared to 0.46) and the FPR has seen an increase (0.23 compared to 0.14). We have not observed a clear benefit to increasing the number of layers in the Ensemble AE architecture. We leave more extensive testing of the autoencoder network hyperparameters, such as non-bottlenecked architecture\citep{Yong2022DoAN} that does not include a bottleneck layer for future work.  

\begin{table}[h]
\begin{center}
\caption{Study on the number of layers in the Ensemble AE architecture}
\begin{tabular}{l|ccc}
\toprule
Method & $F_1$ $\uparrow$ & FPR $\downarrow$ & AUROC $\uparrow$\\
\midrule
Ensemble with 5 layers & 0.43 & 0.23 & 0.74 \\
Ensemble with 3 layers & 0.46 & 0.14 & 0.72 \\
\bottomrule
\end{tabular}
\end{center}
\label{tab:design}
\end{table}

\section{Qualitative Results}

\begin{figure*}[h!]
\centering
\includegraphics[width=0.9\textwidth]{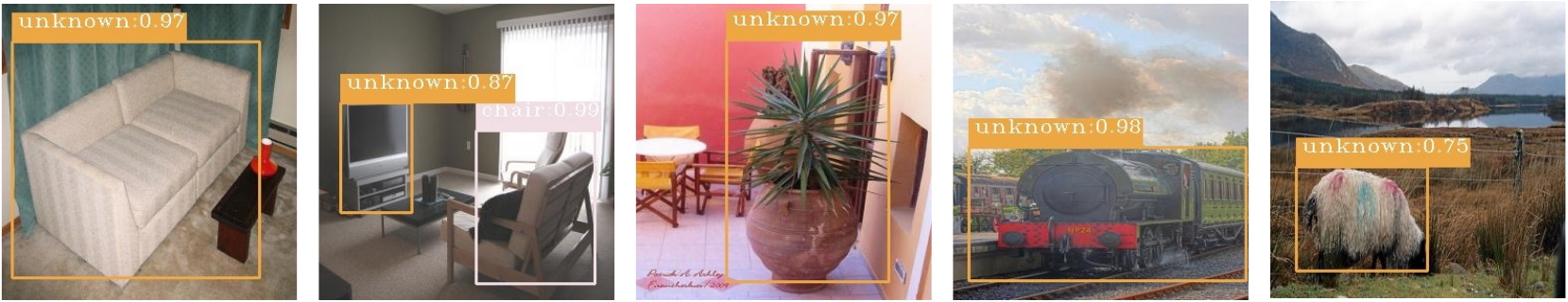} 
\caption{Qualitative results using our OOD Detector and CutLER (Cascade) \citep{wang2023cut} as the open world localization method.}
\label{fig-cascade}
\end{figure*}

\begin{figure*}[h!]
\centering
\includegraphics[width=0.9\textwidth]{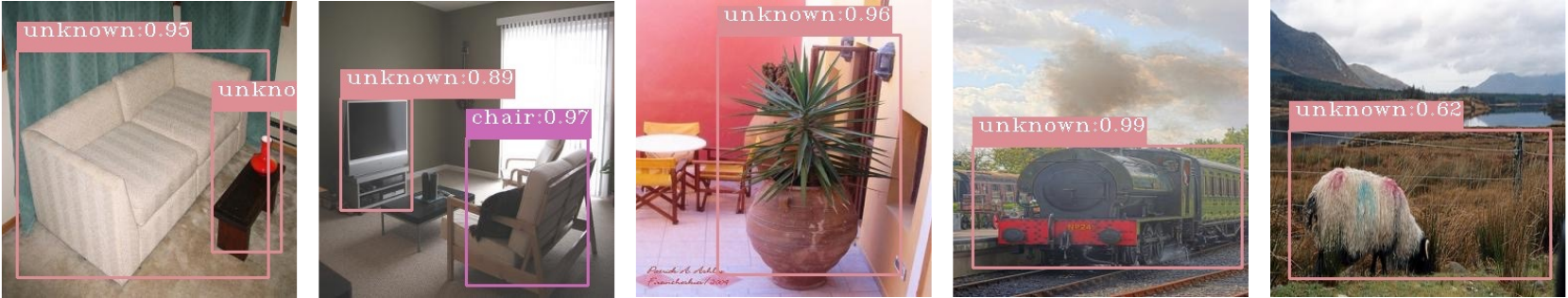} 
\caption{Qualitative results using our OOD Detector and CutLER (MaskRCNN) \citep{wang2023cut} as the open world localization method.}
\label{fig-mrcnn}
\end{figure*}

\begin{figure*}[h!]
\centering
\includegraphics[width=0.9\textwidth]{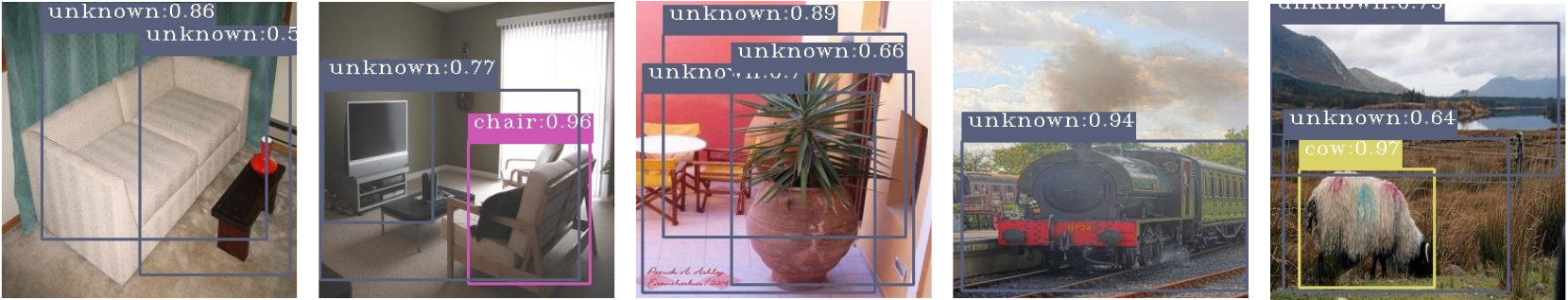} 
\caption{Qualitative results using our OOD Detector and MOST \citep{rambhatla2023most} as the open world localization method.}
\label{fig-most}
\end{figure*}

\begin{figure*}[h!]
\centering
\includegraphics[width=0.9\textwidth]{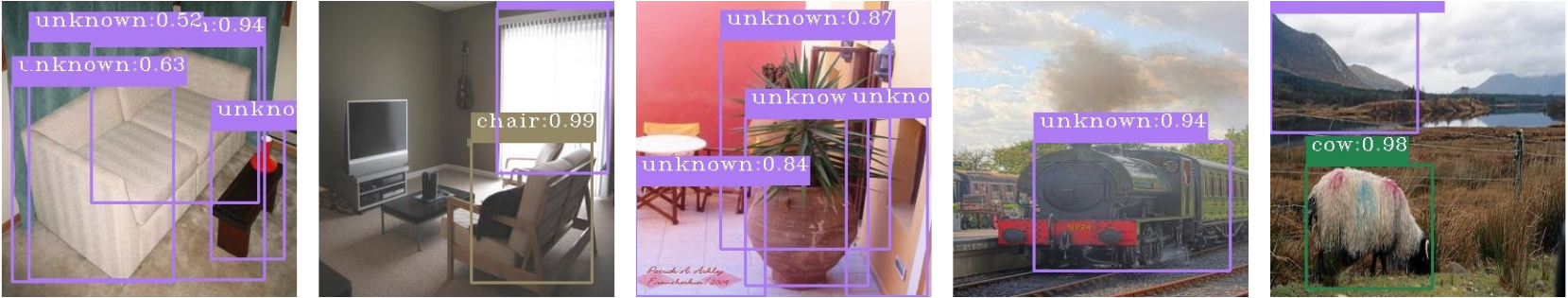} 
\caption{Qualitative results using our OOD Detector and OLN \citep{kim2022learning} as the open world localization method.}
\label{fig-oln}
\end{figure*}

Figures \ref{fig-cascade}, \ref{fig-mrcnn}, \ref{fig-most}, \ref{fig-oln} shows the qualitative results of our approach. We show how different localization methods for OOD classes impact the quality of results. In all cases, the model was trained on 15 ID classes and tested on 20 classes consisting of 15 ID and 5 OOD classes. The OOD classes includes instances of 'potted plant', 'sofa', 'train' and 'tv monitor'. The aim is to detect instances of both ID categories with their class labels and OOD categories with an 'unknown' class label. As seen in the figures, CutLER \citep{wang2023cut} is able to localize OOD objects more effectively compared to OLN \citep{kim2022learning} and MOST \citep{rambhatla2023most}. Both OLN and MOST result in a high number of proposals for OOD objects, some of which are inaccurate or miss relevant objects completely. On the other hand, images with 'sheep' are sometimes misclassified as 'cow', we attribute this to the high degree of similarity between these two classes and similar background that the objects appear in.

